# A Neural Model for User Geolocation and Lexical Dialectology

**Afshin Rahimi**   **Trevor Cohn**   **Timothy Baldwin**
School of Computing and Information Systems
The University of Melbourne
arahimi@student.unimelb.edu.au
{t.cohn,tbaldwin}@unimelb.edu.au

## Abstract

We propose a simple yet effective text-based user geolocation model based on a neural network with one hidden layer, which achieves state of the art performance over three Twitter benchmark geolocation datasets, in addition to producing word and phrase embeddings in the hidden layer that we show to be useful for detecting dialectal terms. As part of our analysis of dialectal terms, we release DAREDS, a dataset for evaluating dialect term detection methods.

## 1 Introduction

Many services such as web search (Leung et al., 2010), recommender systems (Ho et al., 2012), targeted advertising (Lim and Datta, 2013), and rapid disaster response (Ashktorab et al., 2014) rely on the location of users to personalise information and extract actionable knowledge. Explicit user geolocation metadata (e.g. GPS tags, WiFi footprint, IP address) is not usually available to third-party consumers, giving rise to the need for geolocation based on profile data, text content, friendship graphs (Jurgens et al., 2015) or some combination of these (Rahimi et al., 2015b,a). The strong geographical bias, most obviously at the language level (e.g. Finland vs. Japan), and more subtly at the dialect level (e.g. in English used in north-west England vs. north-east USA vs. Texas, USA), clearly reflected in language use in social media services such as Twitter, has been used extensively either for geolocation of users (Eisenstein et al., 2010; Roller et al., 2012; Rout et al., 2013; Han et al., 2014; Wing and Baldridge, 2014) or dialectology (Cook et al., 2014; Eisenstein, 2015). In these methods, a user is often represented by the concatenation of their tweets, and the geolocation model is trained on a very small percentage of explicitly geotagged tweets, noting the potential biases implicit in geotagged tweets (Pavalanathan and Eisenstein, 2015).

Lexical dialectology is (in part) the converse of user geolocation (Eisenstein, 2015): given text associated with a variety of regions, the task is to identify terms that are distinctive of particular regions. The complexity of the task is two-fold: (1) localised named entities (e.g. sporting team names) are not of interest; and (2) without semantic knowledge it is difficult to detect terms that are in general use but have a special meaning in a region.

In this paper we propose a text-based geolocation method based on neural networks. Our contributions are as follows: (1) we achieve state-of-the-art results on benchmark Twitter geolocation datasets; (2) we show that the model is less sensitive to the specific location discretisation method; (3) we release the first broad-coverage dataset for evaluation of lexical dialectology models; (4) we incorporate our text-based model into a network-based model (Rahimi et al., 2015a) and improve the performance utilising both network and text; and (5) we use the model's embeddings for extraction of local terms and show that it outperforms two baselines.

## 2 Related Work

Related work on Twitter user geolocation falls into two categories: text-based and network-based methods. Text-based methods make use of the geographical biases of language use, and network-based methods rely on the geospatial homophily of user–user interactions. In both cases, the assumption is that users who live in the same geographic area share similar features (linguistic or interactional). Three main text-based approaches are: (1) the use of gazetteers (Lieberman et al., 2010; Quercini et al., 2010); (2) unsupervised text clustering based on topic models or similar (Eisenstein et al., 2010; Hong et al., 2012; Ahmed et al.,

2013); and (3) supervised classification (Ding et al., 2000; Backstrom et al., 2008; Cheng et al., 2010; Hecht et al., 2011; Kinsella et al., 2011; Wing and Baldridge, 2011; Han et al., 2012; Rout et al., 2013), which unlike gazetteers can be applied to informal text and compared to topic models, scales better. The classification models often rely on less than 1% of geotagged tweets for supervision and discretise real-valued coordinates into equal-sized grids (Serdyukov et al., 2009), administrative regions (Cheng et al., 2010; Hecht et al., 2011; Kinsella et al., 2011; Han et al., 2012, 2014), or flat (Wing and Baldridge, 2011) or hierarchical $k$-d tree clusters (Wing and Baldridge, 2014). Network-based methods also use either real-valued coordinates (Jurgens et al., 2015) or discretised regions (Rahimi et al., 2015a) as labels, and use label propagation over the interaction graph (e.g. @-mentions). More recent methods have focused on representation learning by using sparse coding (Cha et al., 2015) or neural networks (Liu and Inkpen, 2015), utilising both text and network information (Rahimi et al., 2015a).

Dialect is a variety of language shared by a group of speakers (Wolfram and Schilling, 2015). Our focus here is on geographical dialects which are spoken (and written in social media) by people from particular areas. The traditional approach to dialectology is to find the geographical distribution of known lexical alternatives (e.g. *you*, *yall* and *yinz*: (Labov et al., 2005; Nerbonne et al., 2008; Gonçalves and Sánchez, 2014; Doyle, 2014; Huang et al., 2015; Nguyen and Eisenstein, 2016)), the shortcoming of which is that the alternative lexical variables must be known beforehand. There have also been attempts to automatically identify such words from geotagged documents (Eisenstein et al., 2010; Ahmed et al., 2013; Cook et al., 2014; Eisenstein, 2015). The main idea is to find lexical variables that are disproportionately distributed in different locations either via model-based or statistical methods (Monroe et al., 2008). There is a research gap in evaluating the geolocation models in terms of their usability in retrieving dialect terms given a geographic region.

We use a text-based neural approach trained on geotagged Twitter messages that: (a) given a geographical region, identifies the associated lexical terms; and (b) given a text, predicts its location.

## 3 Data

We use three existing Twitter user geolocation datasets: (1) GEOTEXT (Eisenstein et al., 2010), (2) TWITTER-US (Roller et al., 2012), and (3) TWITTER-WORLD (Han et al., 2012). These datasets have been used widely for training and evaluation of geolocation models. They are all pre-partitioned into training, development and test sets. Each user is represented by the concatenation of their tweets, and labeled with the latitude/longitude of the first collected geotagged tweet in the case of GEOTEXT and TWITTER-US, and the centre of the closest city in the case of TWITTER-WORLD.[1] GEOTEXT and TWITTER-US cover the continental US, and TWITTER-WORLD covers the whole world, with 9k, 449k and 1.3m users, respectively as shown in Figure 1.[2] **DAREDS** is a dialect-term dataset novel to this research, created from the Dictionary of American Regional English (DARE) (Cassidy et al., 1985). DARE consists of dialect regions, their terms and meaning.[3] It is based on dialectal surveys from different regions of the U.S., which are then postprocessed to identify dialect regions and terms. In order to construct a dataset based on DARE, we downloaded the web version of DARE, cleaned it, and removed multi-word expressions and highly-frequent words (any word which occurred in the top 50k most frequent words, based on a word frequency list (Norvig, 2009). For dialect regions that don't correspond to a single state or set of cities (e.g. *South*), we mapped it to the most populous cities within each region. For example, within the *Pacific Northwest* dialect region, we manually extracted the most populous cities (Seattle, Tacoma, Portland, Salem, Eugene) and added those cities to DAREDS as sub-regions.

The resulting dataset (DAREDS) consists of around 4.3k dialect terms from 99 U.S. dialect regions. DAREDS is the largest standardised dialectology dataset.

## 4 Methods

We use a multilayer perceptron (MLP) with one hidden layer as our location classifier, where the input is $l_2$ normalised bag-of-words features for a given user. We exclude @-mentions, words with document frequency less than 10, and stop words. The output is either a $k$-d tree leaf node or $k$-means discretisation of real-valued coordinates of training locations, the output of which is visualised for TWITTER-US in Figure 2. The hidden layer output provides word (and phrase, as bags of words) embeddings for dialectal analysis.

---

[1] The decision as to how a given user is labeled was made by the creators of the original datasets, and has been preserved in this work, despite misgivings about the representativeness of the label for some users.

[2] The datasets can be obtained from https://github.com/utcompling/textgrounder

[3] http://www.daredictionary.com/

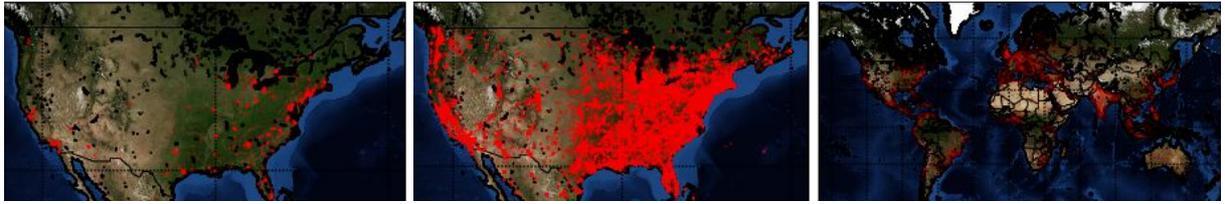

Figure 1: The geographical distribution of training points in GEOTEXT (left), TWITTER-US (middle) and TWITTER-WORLD (right). Each red point is a training user. The number of training users is 5.6K, 429K and 1.3M in each dataset respectively. Despite the number of training instances being higher in TWITTER-WORLD (right) compared to TWITTER-US (middle), there is greater user density in the case of the latter.

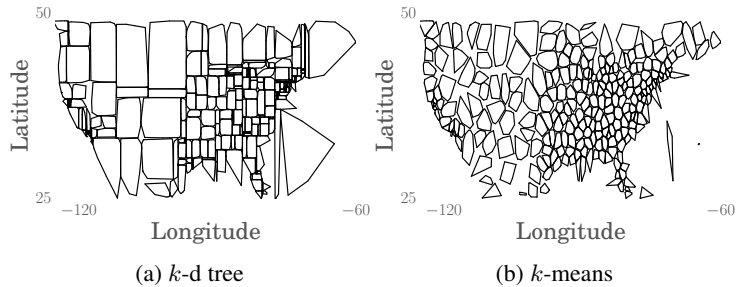

(a) $k$-d tree

(b) $k$-means

Figure 2: Discretisation of TWITTER-US training coordinates using $k$-d tree and $k$-means clustering shown by the convex hulls around the training points within each cluster.

The number of regions, regularisation strength, hidden layer and mini-batch size are tuned over development data and set to (32, $10^{-5}$, 896, 100), (256, $10^{-6}$, 2048, 10000) and (930, $10^{-6}$, 3720, 10000) for GEOTEXT, TWITTER-US and TWITTER-WORLD, respectively. The parameters are optimised using Adamx (Kingma and Ba, 2014) using Lasagne/Theano (Theano Development Team, 2016). Following Cheng (2010) and Eisenstein (2010), we evaluated the geolocation model using mean and median error in km ("Mean" and "Median" resp.) and accuracy within 161km of the actual location ("Acc@161"). Note that lower numbers are better for Mean and Median, and higher numbers better for Acc@161.

[4] The results reported in Rahimi et al. (2015b; 2015a) for TWITTER-WORLD were over a superset of the dataset; the results reported here are based on the actual dataset.

While the focus of this paper is text-based user geolocation, state-of-the-art results for the three datasets have been achieved with hybrid text+network-based models, where the predictions of the text-based model are fed into a mention network as "dongle" nodes to each user node, providing a personalised geolocation prior for each user (Rahimi et al., 2015a). Note that it would, of course, be possible to combine text and network information in a joint deep learning model (Yang et al., 2016; Kipf and Welling, 2016), which we leave to future work (noting that scalability will potentially be a major issue for the larger datasets).

To test the applicability of the model's embeddings in dialectology, we created DAREDS. The output of the hidden layer of the model is used as embeddings for both location names and dialect terms. Given a dialect region name, we retrieve its nearest neighbours in the embedding space, and compare them to dialect terms associated with that location. We also compare the quality of the embeddings with pre-trained word2vec embeddings and the embeddings from the output layer of LR (logistic regression) (Rahimi et al., 2015b) as baselines. Regions in DAREDS can be very broad (e.g. *SouthWest*), meaning that words associated with those locations will be used across a large number of cities contained within that region. We generate a region-level embedding by simply taking the city names associated with the region, and feeding them as BoW input for LR and MLP and averaging their embeddings for word2vec. We evaluate the retrieved terms by computing recall of DAREDS terms existing in TWITTER-US (1071 terms) at $k \in \{0.05\%, 0.1\%, 0.2\%, 0.5\%, 1\%, 2\%, 5\%\}$ of vocabulary size. The code and the DAREDS dataset are available at https://github.com/afshinrahimi/acl2017.

|  | GEOTEXT | | | TWITTER-US | | | TWITTER-WORLD | | |
|---|---|---|---|---|---|---|---|---|---|
|  | Acc@161 | Mean | Median | Acc@161 | Mean | Median | Acc@161 | Mean | Median |
| TEXT-BASED METHODS | | | | | | | | | |
| Proposed method (MLP + $k$-d tree) | 38 | 844 | 389 | 54 | 554 | 120 | 34 | 1456 | 415 |
| Proposed method (MLP + $k$-means) | 40 | 856 | 380 | 55 | 581 | 91 | 36 | 1417 | 373 |
| (Rahimi et al., 2015b) (LR) | 38 | 880 | 397 | 50 | 686 | 159 | 32 | 1724 | 530 |
| (Wing and Baldridge, 2014) (uniform) | — | — | — | 49 | 703 | 170 | 32 | 1714 | 490 |
| (Wing and Baldridge, 2014) ($k$-d tree) | — | — | — | 48 | 686 | 191 | 31 | 1669 | 509 |
| (Melo and Martins, 2015) | — | — | — | — | 702 | 208 | — | 1507 | 502 |
| (Cha et al., 2015) | — | 581 | 425 | — | — | — | — | — | — |
| (Liu and Inkpen, 2015) | — | — | — | — | 733 | 377 | — | — | — |
| NETWORK-BASED METHODS | | | | | | | | | |
| (Rahimi et al., 2015a) MADCEL-W | 58 | 586 | 60 | 54 | 705 | 116 | 45 | 2525 | 279 |
| TEXT+NETWORK-BASED METHODS | | | | | | | | | |
| (Rahimi et al., 2015a) MADCEL-W-LR | **59** | 581 | **57** | 60 | 529 | 78 | **53** | 1403 | 111 |
| MADCEL-W-MLP | **59** | **578** | 61 | **61** | **515** | **77** | **53** | **1280** | **104** |

Table 1: Geolocation results over the three Twitter datasets, based on the text-based MLP with $k$-d tree or $k$-means discretisation and text+network model MADCEL-W-MLP using MLP with $k$-d tree for text-based predictions. We compare with state-of-the-art results for each dataset.[4] "—" signifies that no results were reported for the given metric or dataset.

## 5 Results

### 5.1 Geolocation

The performance of the text-based MLP model with $k$-d tree and $k$-means discretisation over the three datasets is shown in Table 1. The results are also compared with state-of-the-art text-based methods based on a flat (Rahimi et al., 2015b; Cha et al., 2015) or hierarchical (Wing and Baldridge, 2014; Melo and Martins, 2015; Liu and Inkpen, 2015) geospatial representation. Our method outperforms both the flat and hierarchical text-based models by a large margin. Comparing the two discretisation strategies, $k$-means outperforms $k$-d tree by a reasonable margin. We also incorporated the MLP predictions into a network-based model based on the method of Rahimi et al. (2015a), and improved upon their work. We analysed the Median error of MLP ($k$-d tree) over the development users of TWITTER-US in each of the U.S. states as shown in Figure 3. The error is highest in states with lower training coverage (e.g. Maine, Montana, Wisconsin, Iowa and Kansas). We also randomly sampled 50 development samples from the 1000 samples with highest prediction errors to check the biases of the model. Most of the errors are the result of geolocating users from Eastern U.S. in Western U.S. particularly in Los Angeles and San Francisco.

| NYC | LA | Chicago | Philadelphia |
|---|---|---|---|
| manhattan | lapd | chi | septa |
| ny | wiltern | uic | erked |
| soho | ralphs | metra | philly |
| mets | ucla | depaul | phillies |
| nycc | weho | bears | jawn |
| nyu | lausd | chitown | #udproblems |
| #electriczoo | #hollywoodbowl | cta | dickhead |
| yorkers | lmu | bogus | flyers |
| mta | asf | lbs | irked |
| #thingswhitepeopledo | lacma | lbvvs | erkin |

Table 2: Nearest neighbours of place names.

### 5.2 Dialectology

We quantitatively tested the quality of the geographical embeddings by calculating the micro-average recall of the $k$-nearest dialect terms (in terms of the proportion of retrieved dialect terms) given a dialect region, as shown in Figure 4. Recall at 0.5% is about 3.6%, meaning that we were able to retrieve 3.6% of the dialect terms given the dialect region name in the geographical embedding space. The embeddings slightly outperform the output layer of logistic regression (LR) (Rahimi et al., 2015b)

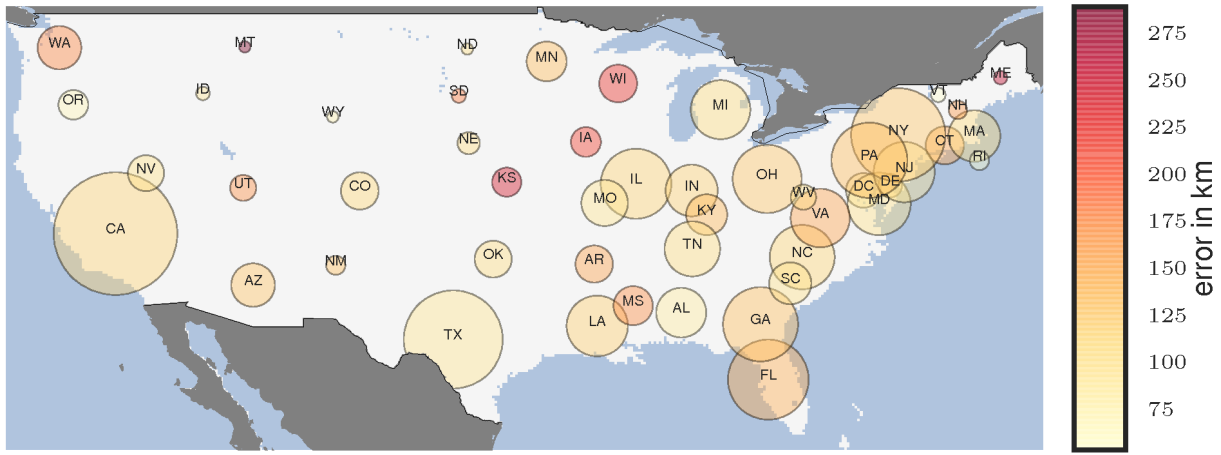

Figure 3: The geographical distribution of **Median** error of `MLP` ($k$-d tree) in each state over the development set of TWITTER-US. The colour indicates error and the size indicates the number of development users within the state.

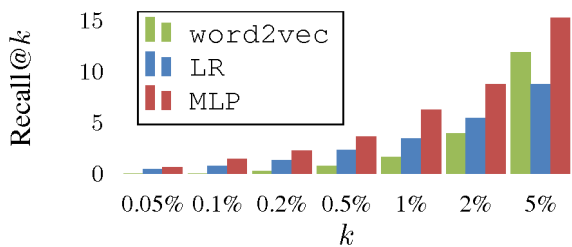

Figure 4: Micro-averaged Recall@$k$ results for retrieving dialect terms given the dialect region name using `LR` embeddings, pre-trained `word2vec` embeddings as baselines, and our embeddings (`MLP`). For example at $k = 0.5\%$ of vocabulary (1292 words), recall is 3.6% for `MLP` compared to 2.3% for `LR` and less than 1% for `word2vec`.

and `word2vec` pre-trained embeddings, but there is still substantial room for improvement.

Our model is slightly better than both baselines, and can retrieve 3.6% of the correct dialect terms given the region name at 0.5% of the total vocabulary, noting the significant performance gap left for future research. It is worth noting that the retrieved terms that are not included in DAREDS are not irrelevant: many of them are toponyms (e.g. city names, rivers, companies, or companies) associated with the given region which are not of interest in dialectology. Equally, some are terms that don't exist in the DARE dictionary but might be of interest for dialectologists because language use in social media is so dynamic that they won't be captured by traditional survey-like approaches. A major shortcoming of this work is that it doesn't incorporate sense distinctions and so can't recover dialect terms that are uniformly distributed but have an idiomatic usage in a particular region.

## 6 Conclusion and Future Work

We proposed a new text geolocation model based on the multilayer perceptron (MLP), and evaluated it over three benchmark Twitter geolocation datasets. We achieved state-of-the-art text-based results over all datasets. We used the parameters of the hidden layer of the neural network as word and phrase embeddings. We performed a nearest neighbour search on a sample of city names and dialect terms, and showed that the embeddings can be used both to discover dialect terms from a geographic area and to find the geographic area a dialect term is spoken. To evaluate the geographical embeddings quantitatively, we created DAREDS, a machine-readable version of the DARE dictionary and compared the performance of dialect term retrieval given dialect region name in terms of recall (Figure 4), and compared the performance to the performance in pre-trained `word2vec` and `LR` embeddings.

## Acknowledgments

We thank the anonymous reviewers for their insightful comments and valuable suggestions. This work was funded in part by the Australian Government Research Training Program Scholarship, and the Australian Research Council.